# UzMorphAnalyser: A Morphological Analysis Model for the Uzbek Language Using Inflectional Endings

Ulugbek Salaev[1, a)]

[1]*Urgench State University*
(*14, Kh.Alimdjan str, Urgench city, 220100, Uzbekistan*)
*0000-0003-3020-7099*
[a)] *Corresponding author: ulugbek.salaev@urdu.uz*

**Abstract.** As Uzbek language is agglutinative, has many morphological features which words formed by combining root and affixes. Affixes play an important role in the morphological analysis of words, by adding additional meanings and grammatical functions to words. Inflectional endings are utilized to express various morphological features within the language. This feature introduces numerous possibilities for word endings, thereby significantly expanding the word vocabulary and exacerbating issues related to data sparsity in statistical models. This paper present modeling of the morphological analysis of Uzbek words, including stemming, lemmatizing, and the extraction of morphological information while considering morpho-phonetic exceptions. Main steps of the model involve developing a complete set of word-ending with assigned morphological information, and additional datasets for morphological analysis. The proposed model was evaluated using a curated test set comprising 5.3K words. Through manual verification of stemming, lemmatizing, and morphological feature corrections carried out by linguistic specialists, it obtained a word-level accuracy of over 91%. The developed tool based on the proposed model is available as a web-based application and an open-source Python library.

## INTRODUCTION

Computational linguistics combines statistical, machine learning, and deep learning models with human natural language modeling using rule-based techniques. The majority of human language technology is challenged by the fact that some languages primarily express grammatical meaning through the use of suffixes and prefixes. In these languages, suffixes and prefixes are more broadly referred to as morphemes, which are the meaningful components of words. A language's morphology refers to both the rules that it uses to assemble morphemes and the actual morphemes that it uses. Morphologically rich languages are those in which morphemes are widely employed in the construction of words. Similar to other Turkic languages, Uzbek has a rich morphology with numerous exceptions in the word structure [1].

Morphology studies how words are built and changed to mean different things, by looking at their parts like stems, prefixes, and suffixes. Inflectional morphology and derivational morphology are the two categories of morphology types. Each of these two groups is significant in different facets of NLP. To express grammatical nuances, an inflectional word ending—one or more suffixes concatenated to the end of a word. These word endings modify only the lexical and grammatical meanings without affecting their core meanings. The goal of the proposed methodology is to develop a morphological analysis model to extract morphological information such as stem, lemma, and morphological features using a complete set of endings (CSE). Also, the model is enriched based on morphological rules and uses CSE to determine the results of morphological analysis in the word. The Uzbek language is considered to be highly inflected, and the large number of suffixes resulting from homonymy causes many problems. According to its word formation structure, a word includes prefixes, suffixes, and base words. The main part of the methodology is developing an inflectional ending dataset tagged with morphological feature tags [2]. This dataset can be used to implement a neural network model for performing the task of morphological analysis on language.

## LITERATURE REVIEW

A number of scientific studies have investigated morphological segmentation methods to reduce corpus size in inflectional languages. The work [3] presented a new morphological segmentation method of Turkic languages based on CSE. This segmentation method was developed in order to divide the word into morphemes in order to reduce the amount of vocabulary in the process of forming word forms and the structure of the corpus.

Due to the inflectional characteristic of Turkic languages, it is very difficult to create a dictionary that covers all the forms of a word. Being able to generate all word forms is a necessary step in the neural machine translation task. The works [3], [4] handled this problem by proposing a new method for breaking down words into smaller parts based on their endings. Using the CSE-based segmentation method for Kazakh, Kyrgyz, and Uzbek languages reduces the size of vocabulary in source texts. Computational experiments in NMT, especially focusing on Kazakh, show promising results. Compared to byte-pair encoding (BPE), the CSE-based method improves the bilingual evaluation understudy score for Kazakh-English and English-Kazakh pairs.

Several models using finite state machines (FSM) have been proposed for morphological analysis in Uzbek language [5]. In [6], authors writers employ automata to extract morphological information from words based on established grammatical rules. However, rather than utilizing grammatical principles, this approach makes use of a database of word forms. Another work [7], uses the FSM model to investigate stemming in the Uzbek language with an emphasis on retrieving pertinent morphological information. For the Uzbek language, researchers have developed morphotactic rules [8] and rule-based morphological analysis models [9]. Uzbek words have a complex structure with many exceptional cases, although in developing rule-based morphological analysis models. Issues such as homonymy, synonymy of affixes, and vowel harmony can lead to errors in morphological segmentation and decrease the accuracy of FSM-based analysis models.

Since the language under the study of this research work is Uzbek, it is noteworthy to include the very recent NLP works in the language, with an effort to reduce its label as a low-resource language. One of those such works include fundamental NLP resources such as stop words dataset [10], [11] , parallel texts datasets for machine translation model creation [12], and many more to come.

## METHODOLOGY

In this section, we describe the detailed information of the proposed methodology for the Uzbek morphological analysis. It uses a comprehensive dataset of word endings along with additional datasets. The model includes rule-based techniques for stemming, lemmatization, and morphological analysis. FIGURE 1 provides an overview of the proposed methodology.

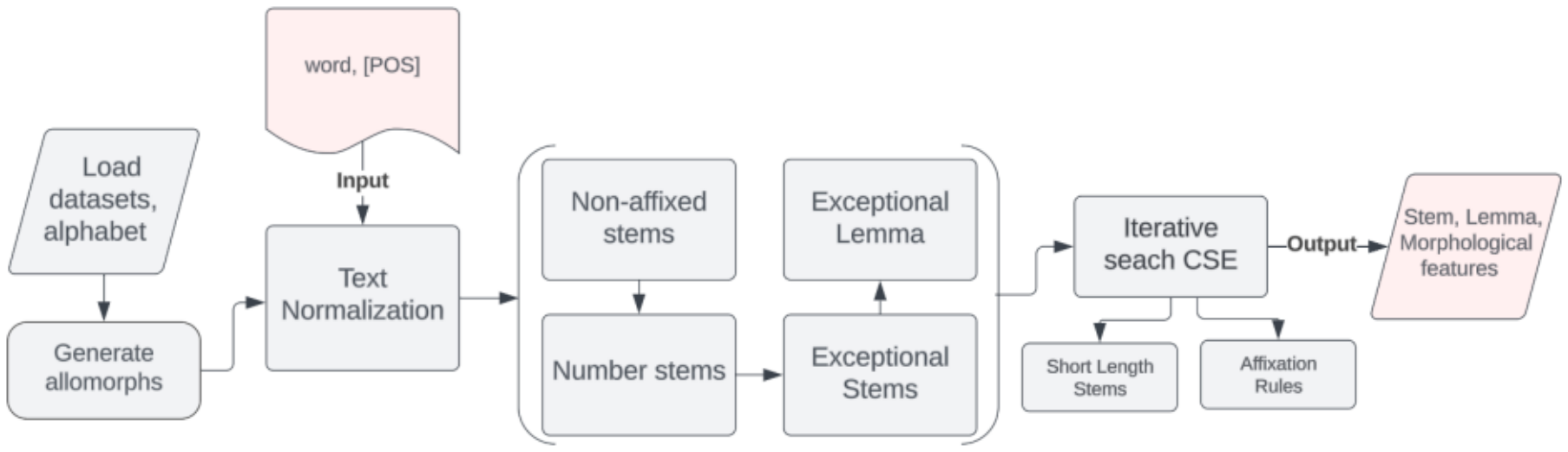

**FIGURE 1.** *Overview of the proposed model.*

The first step of our proposed methodology is to create a comprehensive dataset of word endings for the language. The morphology is divided into two categories: derivational and inflectional morphology. Inflectional morphology includes sequences of suffixes related to both lexical and syntactical aspects. Derivational morphemes change the part of speech of a word. In this study, we focus on maintaining the inflectional part of words. As we gather various forms of words for our dataset, we capture all possible word endings along with their associated morphological information. The morphological information includes a range of linguistic features like part of speech (POS), tense, possession, copula, singular or plural form, question form, case, and others. Selected linguistic experts formed the dataset of word endings to guarantee the accuracy and correctness of this morphological information. By analyzing the morphological structure of words, a database structure is developed to store all the necessary information required for the morphological analysis model. The structure of the database and the relationships between its tables are illustrated in FIGURE 2.

Vowel harmony and affixation rules in Uzbek lead to the presence of multiple forms for a suffix, known as allomorphs. Therefore, in the inflectional ending dataset, we specifically label different variations and allomorphs to account for these variations. To define these allomorphic suffixes, we adopt a notation previously used in the work [13] as following: G:{g,k,q}; K:{g,k}; Q:{g,g‘,k,q}; Y:{a,y}; T:{t,d}; (): the letter within parentheses can be omitted.

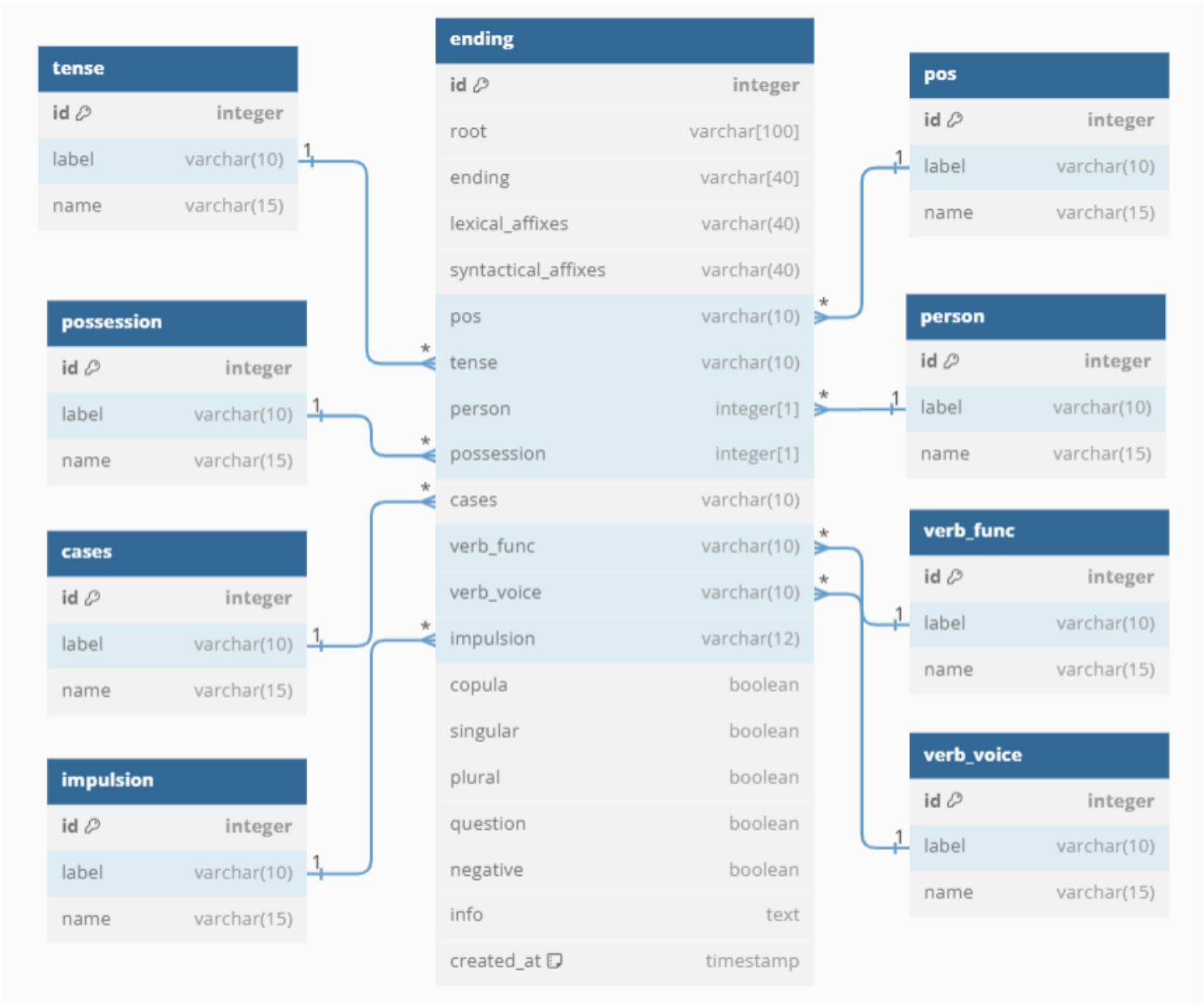

**FIGURE 2.** Structure and Relationship of the Inflectional Ending Dataset.

Based on the structure of this created dataset, a web-based application was developed in order to form the process of collecting inflectional endings and its annotating with morphological features for the Uzbek language. Enrolled linguistic experts are provided with unique usernames and passwords to access the tool to form a dataset. This tool enables users to input new word endings, define their morphological features, and perform checks during the process. The user interface of web-based tool for forming dataset can be seen in FIGURE 3

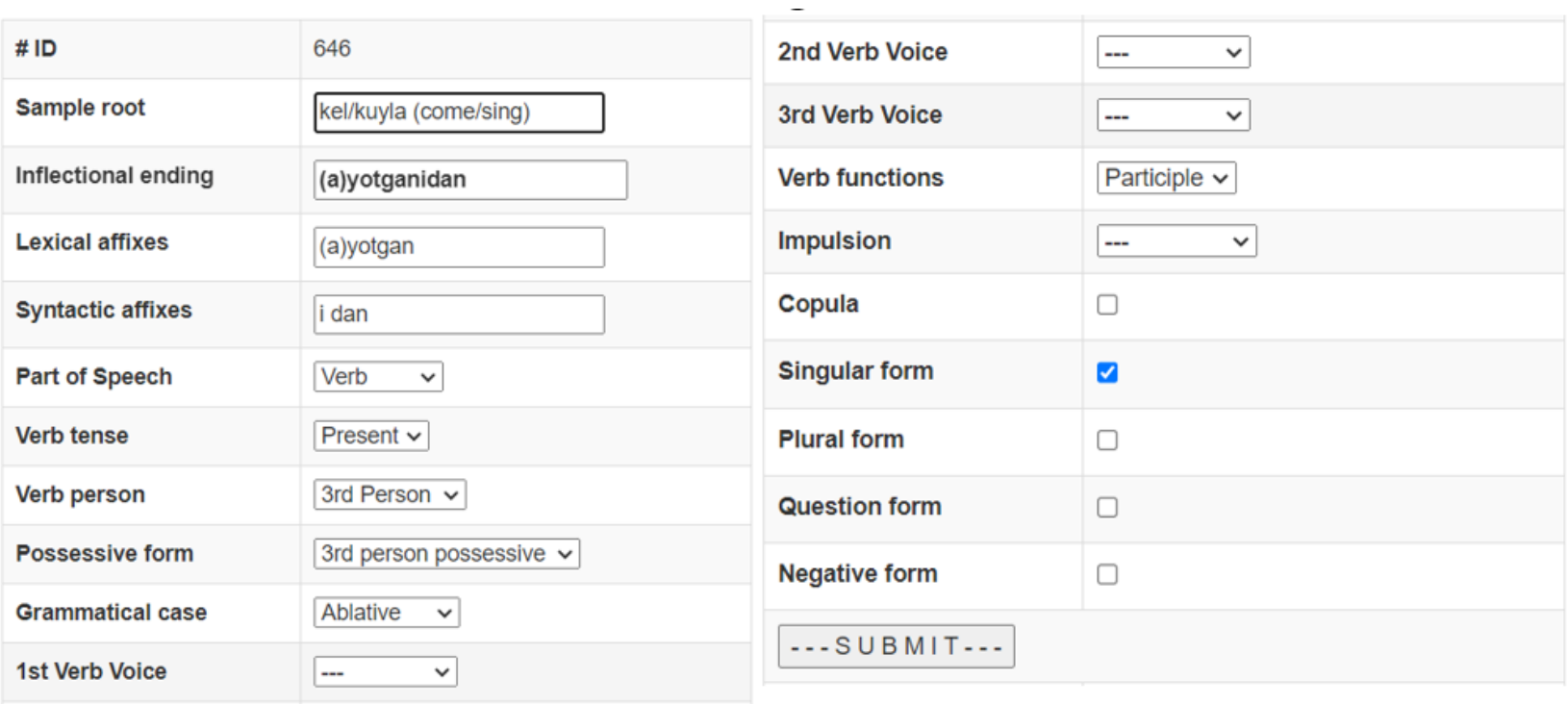

**FIGURE 3.** Web Tool Interface for Inflectional Ending Collection and its Morphological Annotation Process.

A significant portion of inflectional word endings accurately represent the grammatical function of a word as either a verb or a noun. Within the dataset, there are 1205 entries classified as verbs and 150 as nouns. Furthermore, the dataset includes 22 entries for numerals, 10 for adjectives, 20 for pronouns, and 10 entries for adverbs.

In addition to the main dataset of inflectional ending, we have created five additional datasets to enhance the accuracy and performance of the methodology.

1. Exceptional stems [*muzqaymoq (icecream), taqdim (offer), …*], list of stems that may encounter errors when applying rule-based rules.
2. Non-affixed stems [*yoki (or), va (and), …*], list of the stems that never have any affix appended to them.
3. Numbers stems [*uch (three), ming (thousand), …*].
4. List of short length stem (length<=2) [*bu (this), ot (horse), …*].
5. List of mappings for exceptional lemma in the format {word, lemma, inflectional_ending} [{*bitta,bir,ta*} *(one)*, {*singli,singil,i*} *(sister)*].

In the next step of the methodology, all allomorphic options for each item will be generated from the ending column in the main dataset. Subsequently, the modeling process will incorporate alphabet specifications that are used to check morphological rules. Following this, an inputted token will transfer to a text cleaning, which includes converting all characters to lowercase and replacing specific special characters as required. This ensures that the input words are appropriately prepared for further processing and rules checking.

In the following step of the model, an inputted token is checked by using of additional datasets. Based on these additional datasets, the token is extracted to stem, lemma and inflectional ending form, the predicted inflectional ending is verified from the main dataset (inflectional ending dataset). If extracting stem, lemma and inflectional ending do not performed, then all possible options of inflectional endings are generated from the token and each one of them are checked from the inflectional ending dataset. This process performs until obtain stem, lemma and morphological features for the input token.

To enhance the accuracy of morphological analysis, an affixation rules of the language are applied to predicted morpheme segmentation. When the predicted stem is shorter than two characters, an extra verification step is carried out. The short stem dataset is checked to ensure the accuracy of the stem. This validation ensures the correctness of short stems, which improves the reliability of the morphological analysis. These steps produce a list containing the stem, lemma, and morphological information for the provided word and its POS tag. The result will be presented as a human-readable string to clearly representing the analysis.

## RESULT

For evaluation purposes, we analyzed the proposed model using manually selected data from a news corpus, as presented in the works [14], [15], focusing on stemming and lemmatizing tasks. The test data consisted of 40 documents sourced from an open-source corpus, divided into four groups, each containing ten documents. In total, these documents contained 11,952 words, with 5,288 of them being unique words. These unique words account for about 44.24% of all the words in the dataset. In the testing dataset, stem and lemma forms were determined for each item using the spelling dictionary of the Uzbek language [16]. During the evaluation process, the model's performance time averaged about processing 1000 words per second. The model's output was compared with the actual expected output for each word. The outputs were categorized into five distinct cases: one of them is indicating correct predictions while both the stem and lemma forms correctly predicted, while the other four indicated various types of errors encountered during the evaluation. TABLE presents the evaluation results of the model on the compiled corpus, demonstrating the categorization of outputs into different cases. In the table, the surface word, defined as the sample "*daftarimdan*" ("*from my notebook*"), is segmented into three morphemes: *daftar (notebook)* [stem/lemma], -*(i)m* [first person possession], and -*dan* [ablative case].

**TABLE 1.** Evaluation Results of Morphological tasks (Stemming, Lemmatizing)

| Cases | Sample (uz) | Sample (en) | # of token | % |
|---|---|---|---|---|
| Correct prediction | *daftarimdan > daftar* | *from my notebook > notebook* | 4821 | 91.2% |
| Affixes were not removed despite their presence | *daftarimdan > daftarimdan* | *from my notebook > from my notebook* | 24 | 0.5% |
| Affixes were removed from the stem, even though the stem had affixes | *daftarimdan > daft* | *from my notebook > <incorrect word>* | 131 | 2.5% |
| Partial removal of affixes | *daftarimdan> daftarim* | *from my notebook > my notebook* | 158 | 3.0% |
| Affixes were removed from the stem even though there were none | *daftar > daft* | *notebook > <incorrect word>* | 154 | 2.9% |

Most of the errors occur due to the existing of short-length suffix in stems (3rd, 5th cases) and not yet covering all options of CSE in the dataset (4th case).

The Python tool developed for this project is openly accessible and can be easily installed using the popular command within the Python community by the following command: pip install UzMorphAnalyser

To demonstrate the model's performance, a web tool has been developed, an interface of the web tool can be seen in FIGURE 4 To demonstrate the model's performance, a web tool has been developed. The figure shows the interface of the web tool, which presents the input of two tokens {*borgandik (we had gone)*; *maktablarimizni (our schools), NOUN*} and the resulting model analysis. Additionally, a public API system is available for seamlessly integrating the model into other software applications.

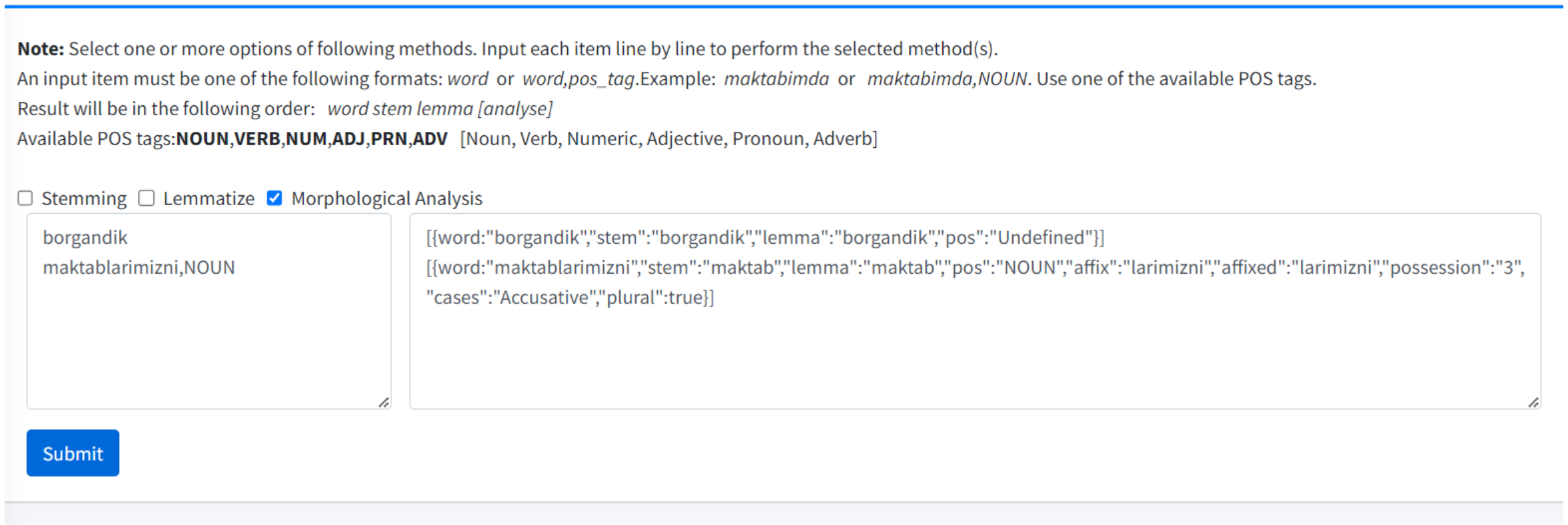

**FIGURE 4.** Web interface of the tool showcasing the performance of the UzMorphAnalyser model

## CONCLUSION

This study conducts a comprehensive morphological analysis model of Uzbek language, covering a range of tasks related to word structure and morphological characteristics. The methodology involves constructing a dataset of inflectional endings, annotated with morphological features by experts in linguistics. Most bound morphemes, especially inflectional endings, are categorized to nouns and verbs, influencing the grammatical attributes of words. The evaluation process encompasses stemming, lemmatization, achieving an impressive accuracy rate of over 91% for stem and lemma validation in a word level. Additionally, we provide Python library, a web interface, and an API to facilitate the utilization of the proposed model.

Our future efforts will focus on improving the output quality of the existing tool by broadening its coverage of inflectional endings, incorporating more morphotactic rules, and integrating a pretrained neural language model. Additionally, we intend to create a comprehensive pipeline capable of handling essential NLP tasks for the Uzbek language, such as morphological generation, POS tagging, and syntactic parsing, in the near future.